\documentclass[letterpaper, 10 pt, conference]{ieeeconf}  

\usepackage{graphicx}
\usepackage[caption=false,font=footnotesize]{subfig}
\usepackage{hyperref}
\usepackage{placeins}
\usepackage{multirow}
\usepackage{dirtytalk}

\IEEEoverridecommandlockouts                              

\newcommand{\ie}{i.\,e.~}

\title{\LARGE \bf
Trust, Shared Understanding and Locus of Control in Mixed-Initiative Robotic Systems
}

\author{Manolis Chiou$^{1}$, Faye McCabe$^{1}$, Markella Grigoriou$^{1}$ and Rustam Stolkin$^{1}$
\thanks{This work was supported by the UKRI-EPSRC grand EP/R02572X/1.}%
\thanks{$^{1}$Extreme Robotics Lab (ERL) and National Center for Nuclear Robotics (NCNR), University of Birmingham, UK}
\thanks       {\tt\small \{m.chiou, fxm493, r.stolkin,\}@bham.ac.uk}%
}

\begin{document}

\maketitle
\thispagestyle{empty}
\pagestyle{empty}

\begin{abstract}

This paper investigates how trust, shared understanding between a human operator and a robot, and the Locus of Control (LoC) personality trait, evolve and affect Human-Robot Interaction (HRI) in mixed-initiative robotic systems. As such systems become more advanced and able to instigate actions alongside human operators, there is a shift from robots being perceived as a tool to being a team-mate. Hence, the team-oriented human factors investigated in this paper (i.e. trust, shared understanding, and LoC) can play a crucial role in efficient HRI. Here, we present the results from an experiment inspired by a disaster response scenario in which operators remotely controlled a mobile robot in navigation tasks, with either human-initiative or mixed-initiative control, switching dynamically between two different levels of autonomy: teleoperation and autonomous navigation. Evidence suggests that operators trusted and developed an understanding of the robotic systems, especially in mixed-initiative control, where trust and understanding increased over time, as operators became more familiar with the system and more capable of performing the task. Lastly, evidence and insights are presented on how LoC affects HRI.

\end{abstract}


\section{Introduction}
As robots are increasingly integrated into human-robot teams, it is pertinent to design human-robot systems that facilitate trust, communication and interdependent working. The irony of automation \cite{Bainbridge1983} is that without careful analysis of how autonomous systems (\ie robots in this context) are introduced and their effect on the work and team, autonomy can reduce situational awareness, decrease attention \cite{Parasuraman2000}, and increase workload, leaving humans to work through tasks too nuanced for automation, in addition to supervision and validation of the autonomy’s actions. 

Having the human-in-the-loop to control, supervise, or even with a peer-to-peer relationship with the robotic system, offers a crucial advantage: the benefits of both agents' capabilities whilst mitigating their weaknesses through collaboration. In remotely-operated robotic systems, robots can keep humans from harm and share workload by conducting tasks remotely and autonomously. Humans can take over in abstract situations requiring contextual, imaginative or moral thinking. Both human-robot team members operate interdependently to solve problems, with work allocation changing dynamically depending on the task. Human operator personality traits, trust in the robot, self-trust, and understanding of the system all can play a role in establishing how the system will be used and relied upon \cite{DeVisser2018}.


This paper focuses on team-oriented human factors aspects of remotely operated Human-Initiative (HI) and Mixed-Initiative (MI) robotic systems, in terms of the operator's trust in the system, shared understanding between the robot and the operator, and the effect the Locus of Control (LoC) personality trait has on HRI. This is done through an empirical study in which human operators conducted a search and rescue navigation task via remotely controlling a realistically simulated mobile robot in either MI or HI control.

The MI and HI systems in our testbed allow for dynamic switching between different Levels of Autonomy (LoA) during task execution. Two different LoAs are used: teleoperation, with an operator manually controlling navigation via a joypad; and autonomy, where the robot autonomously navigates towards human-defined waypoints. In HI control, the human operator is solely responsible for switching LoA based on their judgement \cite{Chiou2016IROS_HI}. In MI control, both the operator and the robot's AI have the authority to initiate actions and switch LoA \cite{Chiou2021_arXiv}, meaning the robot is essentially a peer with equal authority. This makes designing HRI uniquely challenging, as transferring control authority fluidly requires communication, trust and shared understanding within the human-robot team.


This paper contributes evidence that MI robotic systems can be appropriately trusted by operators; trust and shared understanding increase over time; and efficient HRI within the MI system is correlated with an operator's familiarity with the system (i.e. understanding). Additionally, this paper contributes by reporting how LoC affects HRI both in MI and HI control. Evidence suggests that operators with average and high LoC are comfortable giving control of the task to the robot’s AI.




\section{Related Work}

\subsection{Trust in Automation and Human-Robot Interaction}
\label{Section:trust_lit}
According to Chen and Barnes \cite{Chen2014}, the most relevant definition of trust is the one from Lee and See \cite{Lee2004}: Trust is \say{the attitude that an agent will help achieve an individual’s goals in a situation characterized by uncertainty and vulnerability}.  

Hoff and Bashir \cite{Hoff2015} propose three layers of human-automation trust: dispositional, situational, and learned trust. Schaefer et al. \cite{Schaefer2016} categorise factors that affect trust into three categories; human-related, automation-related (the robot in the current context), and environment-related. Hancock et al. \cite{Hancock2011} conducted a meta-analysis of the trust literature and concluded that robot performance-related factors, such as reliability and failure rate, have the largest influence on trust. They argue that more experimental studies on human-related factors are needed. Schaefer et al. \cite{Schaefer2016} meta-analysis on trust further confirms the importance of both automation and human performance-related factors.

Most recent HRI research suggests that individual differences in trust might stem from whether the robot is perceived as an advanced tool, a human-like teammate \cite{Matthews2020}, and/or by cultural differences \cite{Chien2020}. Robot errors are found to have a profound and lasting effect on trust \cite{Wright2020}. Yang et al. \cite{Yang2017} present a data-based model where trust is either exponentially increasing or decreasing based on the robot's reliability and initial user expectations. 

Our literature survey found two studies on trust with control modes similar to HI control, but no studies similar to MI control. In Desai et al., \cite{Desai2012} operators were allowed to switch between fully autonomous and robot-assisted navigation. They found that the effect reduced reliability has on trust becomes less apparent over time. Additionally, when assistance reliability dropped, participants switched to autonomy. Once reliability improved, participants waited longer to switch back to autonomy compared to switching away from autonomy. Nam et al. \cite{Nam2020} proposed computational MDP models to predict an operator's trust levels towards a robotic swarm equipped with a suggestion system regarding LoA switching. They found that the formation shape of the swarm affected trust when task performance was not readily understandable.

In summary, although there is a body of literature on trust in autonomous systems, the field of HRI requires more studies that explore trust in scenarios where operators cooperatively perform tasks using remotely controlled variable-autonomy robotic systems. Often, trust in HRI is studied in tasks with a high level of abstraction or theoretical scenarios \cite{Matthews2020, Wright2020, Wang2016}; when using decision aids or recommendation robots \cite{Yang2017, Wang2016, Nam2020}; or by observing the robotic system conduct a task \cite{Soh2018}. To the best of the authors' knowledge, there is a gap in the literature regarding the empirical investigation of trust in robotic systems where the LoA varies dynamically during task execution. This is especially true for MI robotic systems where the autonomous agent has a peer-to-peer relationship with the operator and can actively take control and change the LoA.

\subsection{Locus of Control} 
\label{section:LoC_lit}
Locus of Control (LoC) is a personality trait and refers to the degree people believe that their actions and behaviours affect outcomes in events in their lives. Simply put, it is the degree of control people believe they have over events. It is usually divided into internal and external LoC. An internal LoC means that a person believes that outcomes and events in their life derive primarily from their own actions and efforts, as they see a causal relationship between their behaviour and rewards. In contrast, an external LoC means that a person believes the outcome of events is mostly affected by forces beyond their control, e.g. luck, fate, powerful others \cite{Rotter1966,Furnham1993}. 

We hypothesise that due to its nature, LoC affects HRI and consequently performance in systems where the operator needs to cooperate with an autonomous agent to control a remote robot to achieve a task. Despite its potential importance, to the best of the authors' knowledge, there is a lack of research considering the operator's LoC personality trait in human-robot teams. In this paper, we aim to address this gap in the literature by investigating the effect of LoC in HI and MI robotic systems.

Our literature survey found two studies in shared control telepresence robots. Takayama et al. \cite{Takayama2011} found that people with an internal LoC took longer to complete an obstacle course than people with an external LoC. Although not statistically significant, they also report evidence that autonomous assistance seemed to degrade performance for operators with a very high internal LoC. This might suggest that individuals with a high internal LoC have problems giving up control to an autonomous system. In Acharya et al. \cite{Acharya2018} average and high internal locus operators have the same level of performance. Additionally, the high internal LoC operators issued the most commands to the robot and had a high percentage of command conflicts. These results suggest this group may seek more control and grow frustrated with the robot’s response.

These studies will be used to inform our discussion better. However, a direct comparison between our work and \cite{Takayama2011, Acharya2018} cannot be made. This is because our work uses MI and HI control and not shared control, meaning the operator interacts with the robotic system differently. In our MI and HI control implementation, the system has the capability to switch between different LoAs on-the-fly (\ie teleoperation and autonomy). In contrast, in shared control, the robot moves according to a scheme for continuously blending control signals from both human and an autonomous agent. Additionally, we use a different LoC scale (i.e. the ICI) as it \cite{Takayama2011} uses the Rotter's scale \cite{Rotter1966} and \cite{Acharya2018} uses an abbreviated version of the LoC Questionnaire (LOCQ). However, all three scales are general measures of LoC. 

\section{Mixed-Initiative and Human-Initiative Control Implementations}

In this paper, we assume a human-robot system that has two LoAs: \textit{teleoperation}, in which the human operator controls the robot manually with the joypad; and \textit{autonomy}, in which the operator clicks on a desired location on the map, and the robot's AI plans and executes a trajectory with the robot to that location. The human-robot system is able to switch between these two LoAs by using two different variable autonomy methods: either Human-Initiative (HI) or Mixed-Initiative (MI) control. 

The HI control switcher used is detailed in \cite{Chiou2016IROS_HI}. The control switcher allows the operator to switch between different LoAs on-the-fly and at any time by pressing a joypad button. It is based on the ability and authority of a human operator to initiate LoA switches based on their own judgment to perform a task efficiently (e.g. improving task performance or overcoming difficult performance degrading situations). The robot's autonomous agent does not have any authority to switch LoAs.

In contrast, MI control is defined in \cite{jiang2015mixed} as \say{a collaboration strategy for human-robot teams where humans and robots opportunistically seize (relinquish) initiative from (to) each other as a mission is being executed...}. In the current context, MI control refers to the authority of both the robot's MI control switcher and the operator to initiate LoA switches at any moment. The MI control switcher used was proposed in \cite{Chiou2021_arXiv} and uses an expert-guided approach to initiate LoA switches. It assumes the existence of an AI task expert (i.e. a navigation planner in this paper) that given a navigational goal, can provide the expected/ideal task performance for the human-robot system in the absence of performance-degrading factors. The comparison between the system's run-time performance with the expected expert performance yields an online task effectiveness metric called "goal-directed motion error". In essence, this is the difference between the robot's current motion (\ie linear velocity) and the motion of the robot required to achieve its goal (\ie reach a target location) according to the expert planner. Hence, this metric expresses how effectively the system performs the navigation task, and the MI control switcher uses it to infer if a LoA switch is needed. In practice, the MI control switcher parameters were trained to use this error to initiate LoA switches based on what the human operators did to improve performance on data from previous experiments. For further details regarding the expert-guided MI control switcher, please refer to \cite{Chiou2021_arXiv}.


\section{Experimental Study}

A disaster response inspired experiment in which a remotely operated mobile robot had to navigate and look for a victim was conducted. The aim was, within the context of HI and MI control, to investigate: (i) how learning (\ie skills acquisition) and training affects task performance; and (ii) the LoC, trust in the robot, self-trust, and shared understanding between the robot and the operator. The reader can refer to \cite{Chiou2019_SMC_learning_effects} for further information on learning and training effects on task performance. In this paper we investigate (ii) from the perspective of the following hypotheses:

\textbf{H1:} Operators trust the MI and the HI robotic systems to assist them and contribute to the task.

\textbf{H2:} Trust in the variable autonomy robotic systems (\ie both in HI and MI), will improve as the operators use the system more.

Additionally, the following exploratory (i.e. more general) hypotheses were investigated:

\textbf{EH1:} Trust in the robot, self-trust, and shared understanding affect HRI in terms of performance, perceived workload, use of autonomy LoA (i.e. how often it is used compared to teleoperation), and LoA switching. 

\textbf{EH2:} Locus of control affects HRI in terms of use of the autonomy LoA, performance, levels of trust, and LoA switching. 

\subsection{Apparatus and software}

  \begin{figure}[t]
    \centering
     \subfloat[\label{fig:ocu}]{%
       \includegraphics[width=0.49\columnwidth]{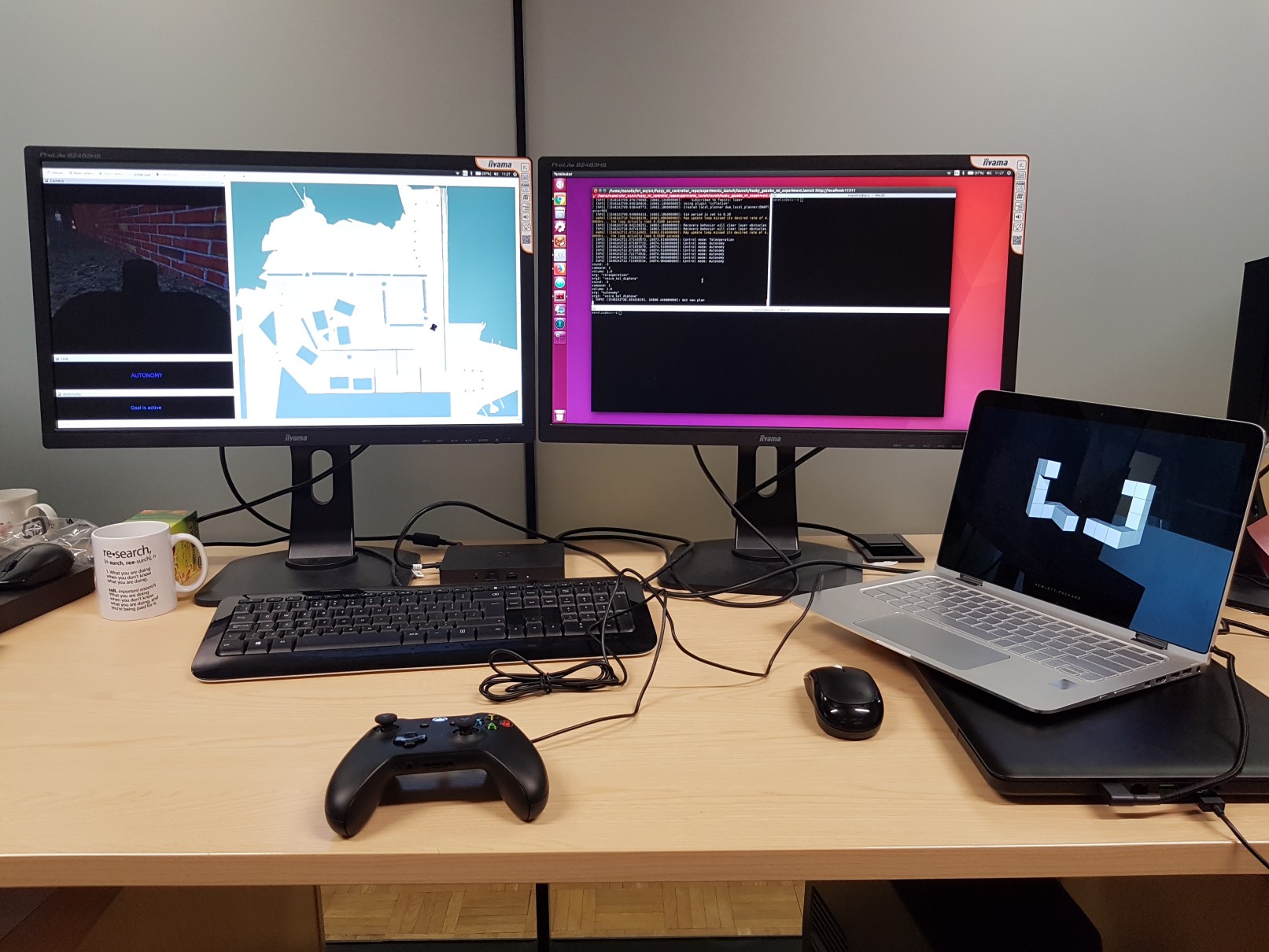}
     }
     \hfill
     \subfloat[\label{fig:secondary_task}]{%
       \includegraphics[width=0.48\columnwidth]{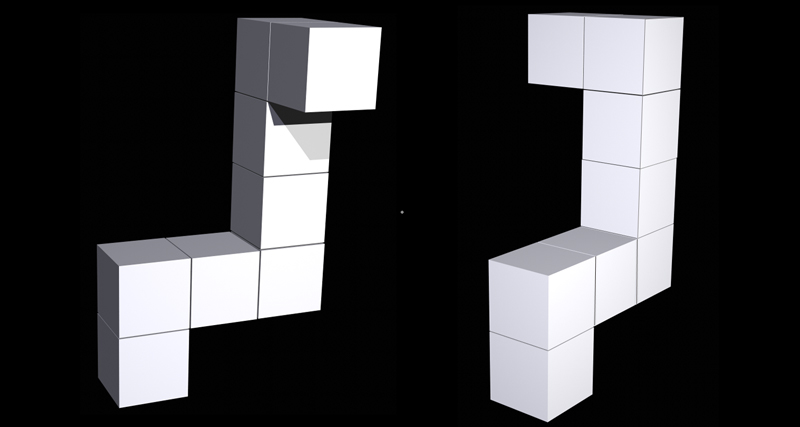}
     }
     \caption{\textbf{\ref{fig:ocu}:} The experimental apparatus: the Operator Control Unit (OCU) composed of a laptop, a joypad, a mouse and a screen showing the GUI; and a laptop presenting the secondary task. \textbf{\ref{fig:secondary_task}:} A typical example of the secondary task \cite{Chiou2019_SMC_learning_effects}.}
     \label{fig:ocu_and_secondary_image}
   \end{figure}

\begin{figure}[thpb]
	\centering
	\includegraphics[width=0.99\columnwidth]{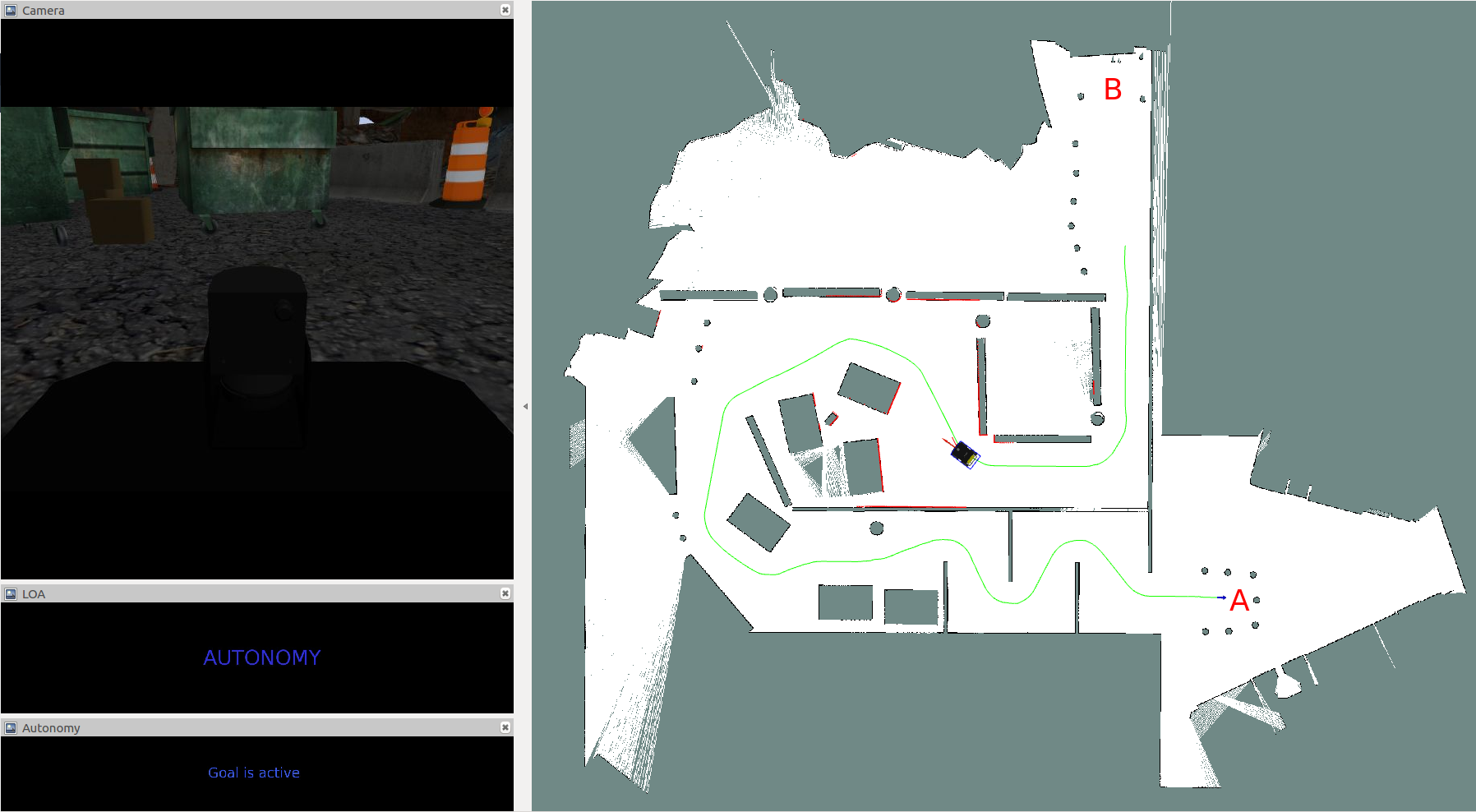}
	\caption{\textbf{Left:} video feed from the camera, the control mode in use and the status of the robot. \textbf{Right:} The map (as created by SLAM) showing the position of the robot, the current goal (blue arrow), the AI planned path (green line), the obstacles’ laser reflections (red) and the walls (black). In the map participants had to navigate from point A to point B and then back again to point A \cite{Chiou2019_SMC_learning_effects}.} 
	\label{fig:gui}
\end{figure}

\begin{figure}[thpb]
	\centering
	\includegraphics[width=0.99\columnwidth]{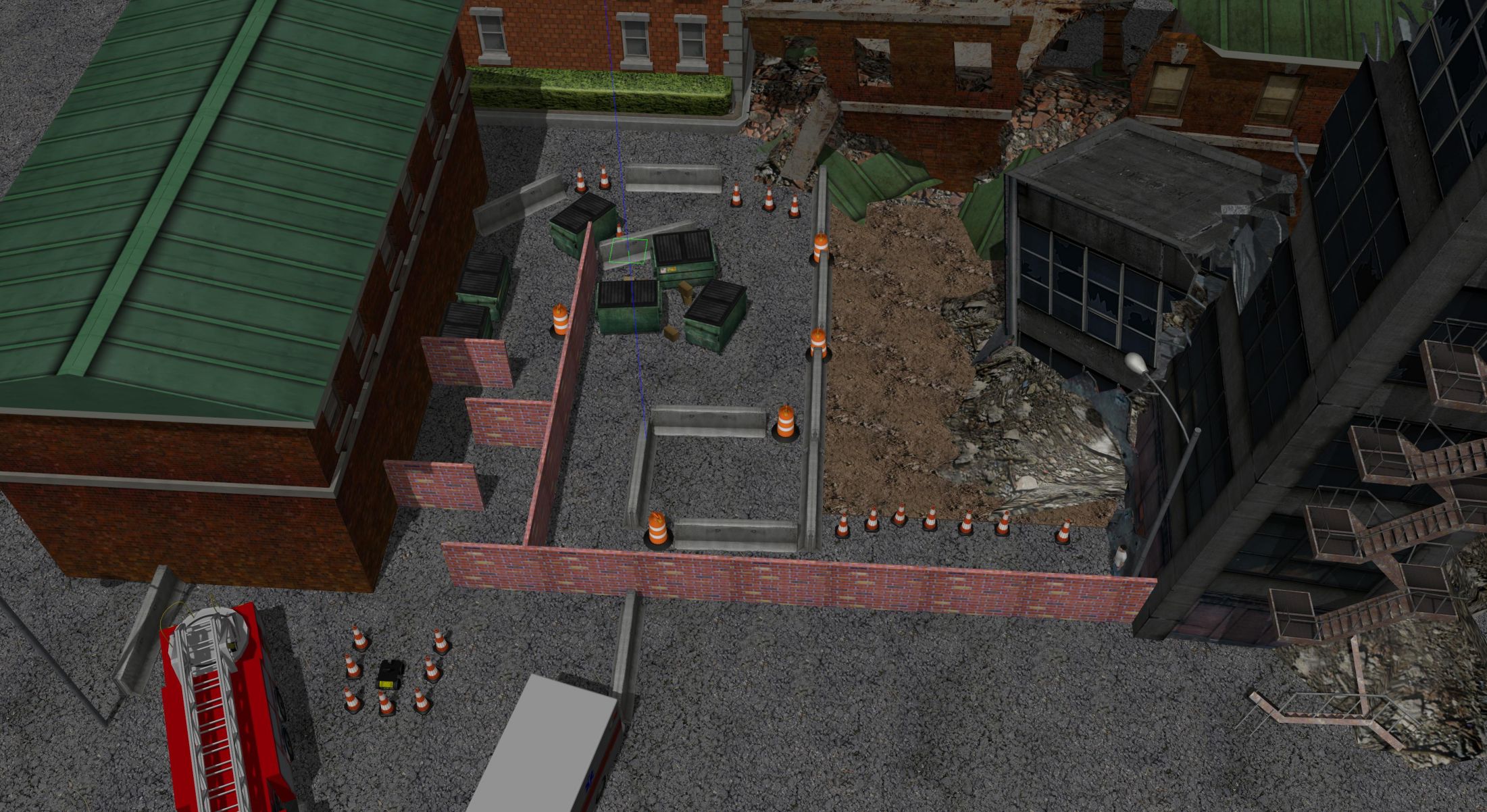}
	\caption{The test arena used, simulating a disaster response scenario \cite{Chiou2019_SMC_learning_effects}.} 
	\label{fig:arena}
\end{figure}

Gazebo, a high-fidelity robotic simulator, was used to simulate the environment and the robotic system (see Figs. \ref{fig:arena} and \ref{fig:gui}). The simulated Husky robot was equipped with a laser range finder and an RGB camera. The simulation was used to improve the repeatability of the experiment.

The software\footnote{The GitHub code repository for running the experiment in ROS is available under MIT license: 
\url{https://github.com/uob-erl/fuzzy_mi_controller}} used was developed in Robot Operating System (ROS) and is described in more detail in \cite{Chiou2016IROS_HI,Chiou2021_arXiv}. The robot's autonomous navigation uses the framework and algorithms (i.e. Dijkstra's algorithm and the dynamic window approach) from ROS navigation stack \cite{Marder-Eppstein2010}.

The robot was controlled via an Operator Control Unit (OCU) (see Fig. \ref{fig:ocu}). The OCU was composed of a mouse and a joypad as input devices, a laptop running the software and a screen showing the Graphical User Interface (GUI) (see Fig. \ref{fig:gui}). 

The experiment's test arena was approximately $576 m^2$ (see Figs. \ref{fig:gui} and \ref{fig:arena}) and the training arena was approximately $72 m^2$. Both arenas had similar level of difficulty. 

The software used to display the secondary task was OpenSesame \cite{Mathot2012OpenSesame}, while the images of the stimuli used were previously created and validated for mental rotation tasks in \cite{Ganis2015}.
   
\subsection{Tasks and Performance Degradation Factors}
 
Participants had to conduct a primary navigation task and a cognitively demanding secondary task. In the primary task, participants had to navigate from point A in Fig. \ref{fig:gui} to point B (where a victim was located) and back to point A.

In the secondary task, taking place parallel to the primary task, participants were presented with a series of images, each showing a pair of 3D objects (see Fig. \ref{fig:secondary_task}). In some of the images, the objects were identical but rotated by 150 degrees. In the rest of the images, the objects were mirror-image objects with opposite chirality. The operator was required to verbally state whether or not the two objects were identical. The images were displayed on the screen of a laptop placed next to the OCU screen (Fig. \ref{fig:ocu}). 

The secondary task represents situations in which the robot operators must interrupt their control of the robot while processing visual or spatial information (e.g. the exact location of a victim or hazard) and relay this information to the rescue team \cite{Murphy2004}.

The secondary task was used to degrade the operator's performance (i.e. teleoperation); artificially generated noise to the laser scanner range measurements was used to degrade autonomous navigation performance (i.e. autonomy). This noise was visible to the operator via the obstacle's laser reflections representation in the GUI (see Fig. \ref{fig:gui}).  In each experimental trial, each of these performance degrading situations occurred once each and at random times. These degrading factors appeared with the restriction that they would not overlap and provide a similar level of degradation in each trial. For every experimental trial, the artificial noise and the secondary task, once initiated, lasted for 30 seconds.

\subsection{Measuring trust, shared understanding and locus of control}
\label{Section:measuring_methods}

Trust and shared understanding were measured by a custom-made 5-point Likert scale questionnaire, with 1 on the scale denoting \say{strongly disagree} and 5 denoting \say{strongly agree}. The questionnaire consisted of 6 questions sub-grouped for measuring three different HRI related aspects. The term team refers to the human-robot team. The questionnaire follows: 

\textbf{Q1:} I trusted the robot to do the right thing at the right time.

\textbf{Q2:} The robot’s performance was an important contribution to the team's success.

\textbf{Q3:} I was confident in my ability to complete the task.

\textbf{Q4:} My performance was an important contribution to the success of the team.

\textbf{Q5:} I accurately perceived the robot's intentions.

\textbf{Q6:} The robot accurately perceived my intentions.

\textit{Trust in the robot} scale questions \textbf{Q1} and \textbf{Q2} measure the operator's trust and confidence in the robot's AI capabilities. These questions were previously validated in \cite{Hoffman2013}.

\textit{Operator's self-trust} scale questions \textbf{Q3} (used from \cite{Nikolaidis2017}) and \textbf{Q4} (adapted from \cite{Hoffman2013}) measure the operator's trust and confidence in their abilities. 

\textit{Shared understanding} scale questions \textbf {Q5} and \textbf {Q6} measure how well the operator thinks the robot understands their intentions and how well the operator understands the robot's actions and intentions. Both questions were adapted from \cite{Hoffman2013}.

The Internal Control Index (ICI) has been used to measure LoC. ICI is a general measure of LoC based on Likert scale questions and was developed by Duttweiler \cite{Duttweiler1984}. The use of ICI in this study is based on three criteria: a) the ICI was developed to counteract some of the original I-E Scale issues \cite{Furnham1993, Duttweiler1984, Meyers1988}; b) it is known to be among the most reliable scales for measuring LoC \cite{Meyers1988, Maltby1996}; c) the ICI is a general measure of LoC with a single total score.

\subsection{Experimental Protocol}
A total of 20 participants took part, with 10 controlling the robot in HI, and the other 10 in MI. Participants were a random sample from the University's student and staff population (University of Birmingham ethical review no.~ERN 19-0199AP2). The participants' age was between 22-41 years old ($M = 30.4, SD = 4.8$), with 13 of them male and 7 female. Before the experiment, participants completed a custom-made 5-point Likert scale questionnaire assessing their level of experience in playing video games, operating robots, and/or related equipment. The number of experienced participants was balanced between the two conditions (\ie between HI and MI). Additionally, participants completed the ICI LoC questionnaire.

Each operator underwent extensive standardised training \cite{Chiou2019_SMC_learning_effects} of approximately 20-25 minutes before the experiment, conducted in a training arena that differed from the experiment arena. Each system and experimental aspect was introduced gradually, and participants were allowed to practice using them: \ie the GUI, the different LoAs, how distances from obstacles in the virtual environment mapped to the robot's pose and movement, and the performance degrading factors. Participants practised using either the MI or HI capabilities to switch between LoAs depending on the experimental condition they were allocated. To ensure that all participants had attained a minimum skill level in controlling the robot, they were not allowed to proceed with the experimental trials until they had first demonstrated that they could complete a training obstacle course within a specific time limit, with no collisions, and while presented with the two performance degrading factors. 

During the experiment, each group of participants performed 5 identical trials of the same tasks. At the end of the third trial (\ie half way through the experiment), all participants had a 3-minute break to reduce the effect of fatigue. At the end of each trial, participants completed the trust and shared understanding questionnaire.

Participants were instructed to perform the primary task (controlling the robot in the navigation task) as quickly and safely (\ie minimising collisions) as possible. Additionally, they were instructed that when presented with the secondary task, they should do it as quickly and as accurately as possible (\ie to have as many correct answers as possible in the 30 second time limit). They were explicitly told that they should prioritise the secondary task over the primary task and should only perform the primary task if the workload allowed.

The participants could only acquire situation awareness (SA) via the GUI (see Fig. \ref{fig:gui}). When a LoA switch occurred from either the operator or the MI control switcher, the operators were alerted in three different ways: a) by an alarm sound identical to the one denoting autopilot disconnection in aircraft; b) by synthetic speech expressing the LoA the system switched to; c) by a GUI notification.

At the end of the experiment, participants completed a raw NASA-TLX task workload questionnaire \cite{Sharek2011}. 

\section{Results and Statistical Analysis}

When comparing the data from the HI and MI conditions, these data were treated as between groups, \ie independent samples, with pairwise comparisons conducted with Mann-Whitney U test. When analysing data between different trials of the same condition (\ie HI or MI), the data were treated as within-subject/repeated measures. The Wilcoxon signed ranks test was used for within-subject pairwise comparisons and the Spearman's rank for correlations. All tests were two-tailed. We consider a result to be statistically significant when it yields a $p$ value less than $0.05$. The trial number is abbreviated as \textit{t1} for the first trial, \textit{t2} for the second, and so on. For correlations, the metrics were averaged across trials, as is standard practice.

To facilitate some comparison with relevant literature that follows similar grouping \cite{Acharya2018}, the ICI scale was divided proportionally in three categories: operators with an ICI score of a) between $28-65$ have a high external LOC; b) between $66-102$ have an average LOC; c) between $103-140$ have high internal LOC. The minimum possible score of the scale is $28$ and the maximum possible score is $140$. In our sample of participants the lowest ICI score was $96$ and the highest $127$. Thus, all participants fell into the average LOC (4 participants) and high internal (16 participants) categories. The average ICI score of participants in MI and HI (see Table \ref{table:results}) did not show a statistical difference according to a Mann-Whitney U test. This means that our sample (i.e. participants) is equally distributed in terms of ICI between conditions (\ie HI and MI). Factors that seem to impact LoC, such as social environment and family structure \cite{Twenge2004} are very hard to control and are outside of the scope of this paper.
 
The following metrics were analysed in conjunction with trust and LoC: primary task completion time (sec); secondary task number of correct answers; secondary task accuracy (\ie the percentage of correct answers from the total responses given); the number of LoA switches; percentage of time spent in autonomy LoA; NASA-TLX score. 

 
\textbf{Results H1:} The mean trust in the robot (i.e. \textbf{Q1} and \textbf{Q2}) across trials was \textit{$M = 4.39, SD = 0.67$} for HI and \textit{$M = 4.71, SD = 0.26$} for MI. The mean self-trust (i.e. \textbf{Q3} and \textbf{Q4}) across trials was \textit{$M = 4.44, SD = 0.56$} for HI and \textit{$M = 4.22, SD = 0.43$} for MI. The mean shared understanding (i.e. \textbf{Q5} and \textbf{Q6}) across trials was \textit{$M = 4.44, SD = 0.61$} for HI and \textit{$M = 4.59, SD = 0.28$} for MI. No statistically significant differences were found in self-trust, trust in robot and shared understanding between HI and MI.

\begin{figure*}[t]
     \subfloat[\label{fig:trust_hi}]{%
       \includegraphics[width=0.9\columnwidth]{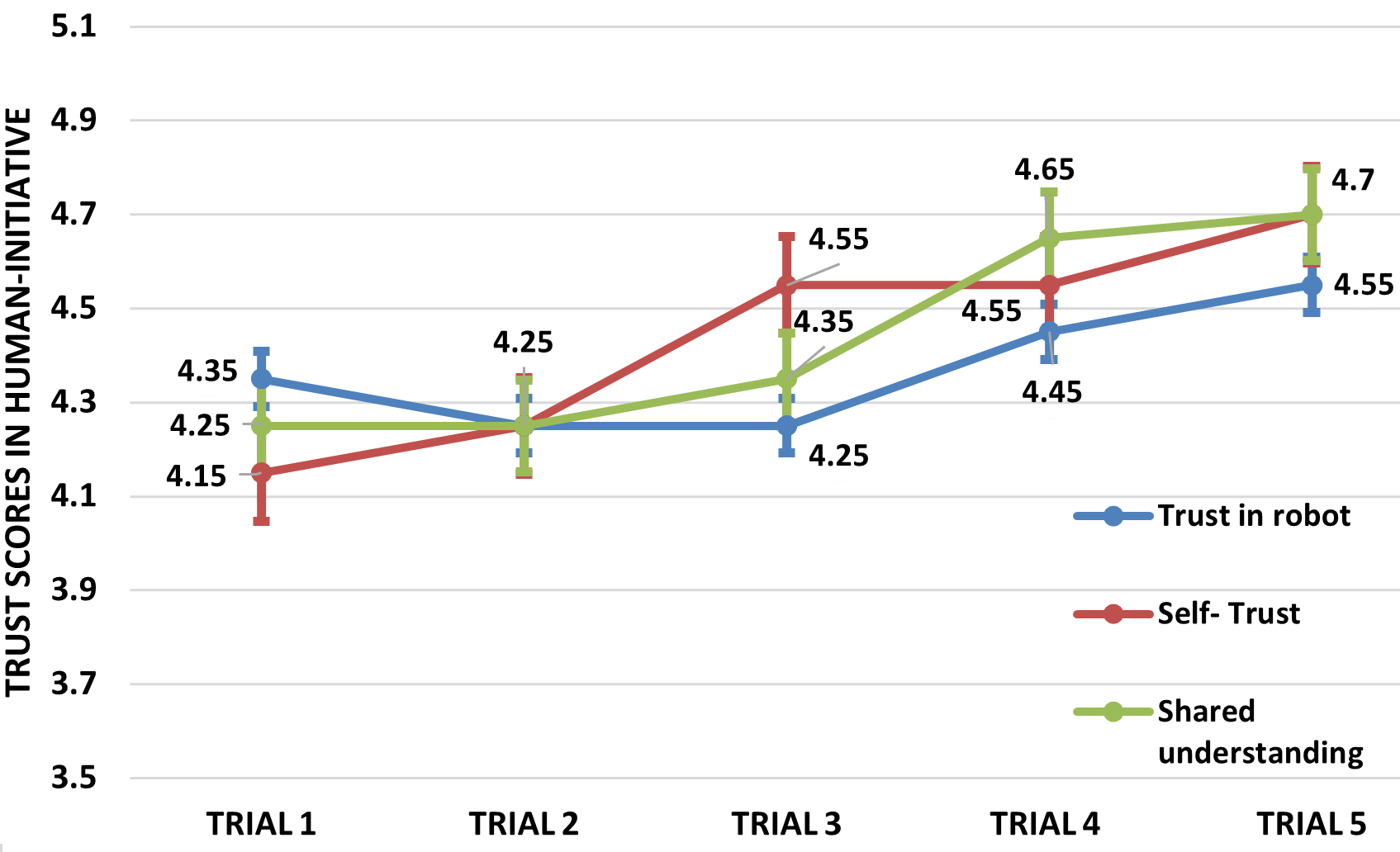}
     }
     \hfill
     \subfloat[\label{fig:trust_mi}]{%
       \includegraphics[width=0.9\columnwidth]{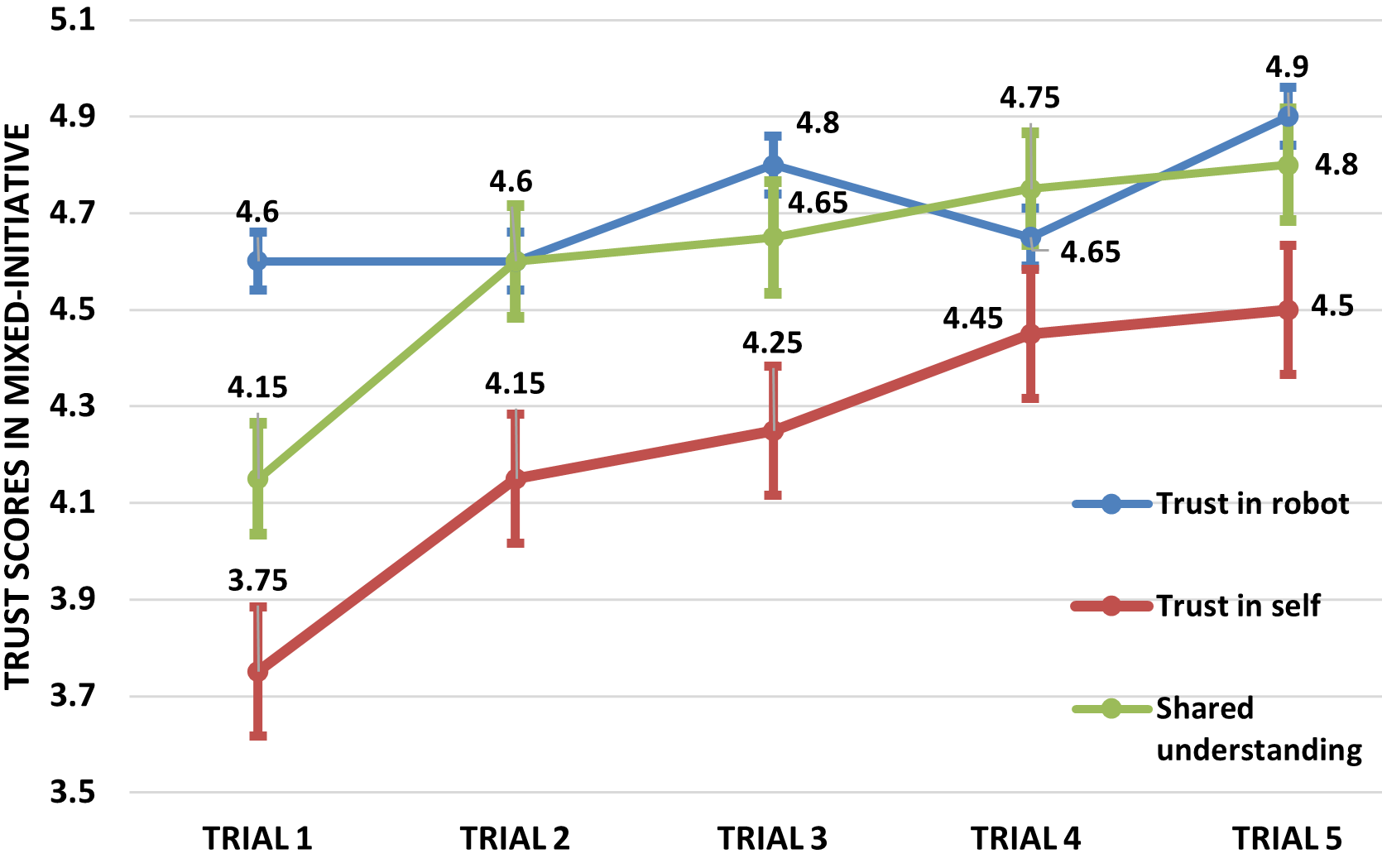}
     }
     \caption{Mean trust and shared understanding scores across trials for \textbf{\ref{fig:trust_hi}} Human-Initiative and  \textbf{\ref{fig:trust_mi}} Mixed-Initiative. The error bars indicate the standard error. The range of values in the y-axis in both graphs is [1,5] (i.e. a 5-point Likert scale).}
     \label{fig:trust}
   \end{figure*}

 \textbf{Results H2:} In HI, no significant difference was found between t1 and t5 concerning trust in the robot, self-trust and shared understanding (see Fig. \ref{fig:trust} and Table \ref{table:results}). In contrast, in MI, trust in the robot (\textit{$Z = -2.21, p = .034$}), self-trust (\textit{$Z = -2.59, p = .01$}) and shared understanding (\textit{$Z = -2.56, p = .01$}) are significantly higher in t5 compared to t1 (see Table \ref{table:results}). As can be seen in Fig \ref{fig:trust} there is a clear increase of trust, self-trust, and shared understanding in MI over time.

\textbf{Results EH1:} In MI, a significant negative correlation was found between shared understanding and operator-initiated LoA switches \textit{$(\rho = -.66, p < .038)$}. A significant negative correlation was found between shared understanding and primary task completion time \textit{$(\rho = -.72, p < .019)$}. No other correlation was found in MI between the trust metrics and any of the other metrics.

In HI, a strong negative correlation was found between self-trust and the NASA-TLX \textit{$(\rho = -0.79, p = 0.006)$}. No other correlation was found between trust in robot, shared understanding and any other metric.
 
\textbf{Results EH2:} In HI, a significant negative correlation \textit{$(\rho = -.67, p < .034)$} between ICI and trust in the robot was found. No significant correlation was found between ICI and other measured metrics. 

In MI, no significant correlation was found between ICI and any of the metrics.

\begin{table}[ht]
\centering
\caption{Descriptive statistics and pairwise comparisons.}
\begin{tabular}{lll}
                                    &                                 &                               \\
\textbf{metric \& condition}        & \textbf{descriptive statistics} & \textbf{significance}         \\ 
\hline
\textbf{Mixed-Initiative:}          &                                 &                               \\
trust in robot t1                   & $M = 4.6 , SD = 0.32$           & \multirow{2}{*}{$p = 0.034$}  \\
trust in robot t5                   & $M = 4.9 , SD = 0.21$           &                               \\ 
\hline
self-trust t1~                      & $M = 3.75 , SD = 0.8$           & \multirow{2}{*}{$p = 0.01$}   \\
self-trust t5                       & $M = 4.5 , SD = 0.53$           &                               \\ 
\hline
shared understanding t1             & $M = 4.15 , SD = 0.47$          & \multirow{2}{*}{$p = 0.01$}   \\
shared understanding t5             & $M = 4.8 , SD = 0.35$           &                               \\ 
\hline
ICI                                 & $M = 111.3 , SD = 10.42$        &                               \\ 
\hline
\textbf{\textbf{Human-Initiative:}} &                                 &                               \\
trust in robot t1                   & $M = 4.35 , SD = 0.78$          & \multirow{2}{*}{$p > 0.05$}   \\
trust in robot t5                   & $M = 4.55 , SD = 0.21$          &                               \\ 
\hline
self-trust t1~ ~                    & $M = 4.15 , SD = 1.13$          & \multirow{2}{*}{$p > 0.05$}   \\
self-trust t5~~                     & $M = 4.7 , SD = 0.26$           &                               \\ 
\hline
shared understanding t1             & $M = 4.25 , SD = 1.11$          & \multirow{2}{*}{$p > 0.05$}   \\
shared understanding t5             & $M = 4.7 , SD = 0.53$           &                               \\ 
\hline
ICI                                 & $M = 108.7 , SD = 6.1$          &                              
\end{tabular}
\label{table:results}
\end{table}


\section{Discussion}

\subsection{Trust and shared understanding}
Regarding \textbf{H1}, the results show that operators trusted the robot's capabilities to contribute to the task both in HI and in MI, hence confirming \textbf{H1}. More specifically, in MI, the evidence shows strong levels of trust denoted by the high proportion of \say{strongly agree} responses, especially in t5. Additionally, operators had confidence in their ability to complete the tasks and had a good shared understanding with the robot, both in HI and in MI. 

In HI, \textbf{H2} evidence suggests that trust in the robot, trust in self, and shared understanding did not change significantly over time, although some non-significant increase can be observed (see Fig. \ref{fig:trust_hi}). This might suggest that the standardised training has provided participants with a close to maximum trust and understanding they could have for the HI system in the context of this experiment. However, the lack of statistically significant evidence could also be because of the relatively small sample size.

In MI, and in contrast with HI, the evidence presented confirms \textbf{H2} and shows that trust and understanding of the system improved over time. This is important evidence pointing towards two directions: reliability of and familiarity with the system. First, in conjunction with the performance increase over time in \cite{Chiou2019_SMC_learning_effects}, this gives evidence to support the reliability of the MI system in the context of this experiment, backed up by the trust model of Yang et al. \cite{Yang2017} which predicts that trust in a robot increases with robot reliability. Additionally, this further contributes to the current literature (see Section \ref{Section:trust_lit}) in showing that trust is mainly driven by performance (e.g. reliability). Secondly, this increase in trust and shared understanding may be related to the increased amount of HRI which occurs in MI compared with HI. The interaction in the MI condition is more complex and meaningful when compared to HI, as the robot's MI control capabilities are more actively involved and perceivable by the operator via the control-switching initiative. This in turn, could have led to increased familiarity with the system over time. An explanation can be found in \cite{Lee2004}: for trust to grow, humans must be familiar with the system’s operations, methods, and constraints. By understanding the “three Ps” of a system \cite{Lee2004, Chancey2017b}; performance, process, and purpose; an operator can develop an appropriate level of trust in a system. Thirdly, a \say{dip} in trust was observed in t4 as three participants rated their trust lower in t4 compared to t3. However, in t5, trust increased again to the highest level across all trials. The authors predict this could have been due to a few negative interactions by chance in t4. The dip in trust was rapidly reconciled in t5.

Regarding \textbf{EH1} in MI, shared understanding was negatively correlated with both the number of operator-initiated LoA switches and the primary task completion time (shorter time means better performance). These results provide evidence that the more an operator understood the MI system, the less they initiated LoA switches and the better they performed. The above evidence explains and adds to the findings in \cite{Chiou2019_SMC_learning_effects} in which the total number of LoA switches was decreasing over trials (i.e. time). Additionally, the performance improvement over time in \cite{Chiou2019_SMC_learning_effects} suggests that operators become more efficient in LoA switching as they use and understand the system more. Lastly, in MI, the lack of correlation with the use of autonomy LoA might suggest that trust mostly reflects the LoA switching capabilities of the MI control switcher, or the system as a whole, rather than explicitly the autonomy LoA. However, trust and the use of autonomy in both HI and MI conditions were relatively high across participants, and hence a bigger and more diverse sample is required to explore this further.

 Regarding \textbf{EH1} in HI, the negative correlation found between self-trust and the perceived workload (i.e. NASA-TLX) means that the more operators trusted themselves and were confident about operating the robot, the lower their perceived workload. This is possibly due to an absence of assistance from the robot in terms of LoA switching. This is further supported by the lack of evidence of such a correlation in MI, in which the robot's MI control switcher was actively helping with LoA switching and the operators with low self-trust did not experience an increase in workload. This could be because the robot takes some of the burden of control and/or psychologically the active AI role makes operators perceive their workload as less intensive, knowing that the robot's autonomous agent will take the initiative if they do not perform well. However, the causality can be the other way around as the self-perceived ability to use the robot's capabilities can be possibly influenced by workload. 

\subsection{Locus of control} 
The negative correlation in the HI condition between ICI and trust in the robot means operators with highly internal LoC trust the robot less. This differs in the MI condition, where no correlation was found. An explanation for this may be that individuals who score highly on internal LoC consider themselves solely responsible, negatively impacting how likely they are to trust somebody else to accomplish the task and transfer control and therefore, responsibility. 

Additionally, an autonomous system that can take control as in MI, may be more likely to be perceived as an assistant or collaborator rather than a tool \cite{DeVisser2018}. In contrast, in HI control, the robot is more likely to be perceived as a tool, as it is unable to interact with the operator actively and must obey their instructions. With no way to experience the more “intelligent” side of the robot, allowing it to assess human performance and take control when necessary, an operator must actively decide to relinquish control. If the operator has a high internal LoC and no way to become familiar with the robot's reliability, they lack the ability to develop more trust in the robot. This is because trust is based upon familiarity with the robot’s behaviour and this would only be possible through relinquishing control, which goes against their internal LoC.

The absence of correlation evidence between primary task performance and the LoC, both in HI and in MI, is in accordance with the findings of \cite{Acharya2018}, that performance between average and high internal LoC operators are indistinguishable. This makes sense as all of our participants fell into the category of having average or high internal locus of control. Hence, \textbf{EH2} in terms of performance cannot be confirmed in the current study. However, it cannot be rejected either due to the lack of high external LoC participants in our sample, requiring further investigation.

Our results did not find evidence of a correlation between ICI and the number of operator-initiated LoA switches in both MI and HI. This might suggest that LoC is not interacting with the number of commands (i.e. LoA switches in our case) the operators give to the robot. This is contrary to the increased shared control commands given to the robot in \cite{Acharya2018} by the high internal locus operators. 

The lack of evidence of correlation both in MI and HI between ICI and the percentage of task time spent in autonomy LoA, in conjunction with the high percentage of time spent in autonomy LoA by the participants, \textit{$M = 79.9, SD = 19.56$} in HI and \textit{$M = 76.8, SD = 18.5$} in MI, might suggest that average and high internal individuals have no issues with giving control to the autonomous navigation in the context of this experiment. This does not agree with the suggestion made by Takayama et al. \cite{Takayama2011} that high internal individuals have problems giving control. However, as mentioned in the related work Section \ref{section:LoC_lit}, a direct comparison with \cite{Acharya2018, Takayama2011} cannot be made.

\subsection{Limitations, insights, and future challenges}

This study aimed to be a starting point towards investigating trust, shared understanding, and locus of control in dynamic variable autonomy systems such as HI and MI. Towards this end, in this section, we will provide insights, discuss how to overcome some of the limitations, and provide suggestions for future research. 

First, although we consider the number of participants sufficient to capture some of the trends accurately (e.g. trust increasing in MI over time) it might have prevented us from finding others (e.g. slower increase in trust in HI). Additionally, correlation does not necessarily mean causation. Establishing causation in human factors research can be challenging. Future experiments should include a large number of participants from different age groups and explicitly control for trust, predisposition to robot use, and personality traits so causal relationships can be investigated.

Second, due to the complex nature of HRI in MI, there is a need for an MI specialised framework of trust able to capture its multiple dimensions. For example, the current trust measurement does not explicitly differentiate between trust in the robot's autonomous navigational ability and trust in the robot's LoA switching capabilities in MI. Future work can tackle this by measuring trust based on momentary interactions (e.g. before and after an MI control switcher initiated changes) and investigating the impact of varying the robot’s navigational reliability on the frequency of LoA switching. Additionally, multidimensional post hoc questionnaires measuring trust in the context of MI should be developed. Once such a framework is validated, it can lead to computational models of trust such as in \cite{Chen2018} which can inform the MI system's LoA switching policies. We consider this a major challenge for the field.

Third, there was no explicit effort to convey trust or make the MI system more transparent to the operator but only follow basic guidelines. Hence, transparency and other factors that affect trust were not explicitly controlled or taken into account. Extensive training mitigated some of these problems by explaining how the system works to the operators and providing them with a basic understanding of the MI algorithm. Future work should explicitly investigate and design interfaces that promote transparency in the MI context. This is a crucial step towards true human-robot teaming. 

Lastly, similar to other related work, all participants had an average or high internal LoC. This perhaps limits the findings only to those users, excluding those with a high external LoC. A future study should aim to incorporate individuals with a larger variety of LoC, although this is difficult to control for. Another way this could be explored is by studying the LoC of real robot operators, the expectation being they would possess a high level of internal LoC, especially concerning the robot's control.

\section{Conclusion}

This paper presented an empirical investigation of HRI in MI and HI variable autonomy systems in terms of trust in the robot, self-trust, shared understanding between the robot and the operator, and the locus of control personality trait. The evidence presented supports the idea that operators learn to trust (i.e. trust increasing over time) the MI robotic system, and their understanding of the system improves over the course of several interactions in the context of this work. Furthermore, evidence was reported on how LoC affects HRI. Average and high internal LoC operators were found to be comfortable giving control to the robot's AI for conducting the navigation task.

Lastly, this work provided insights and highlighted research areas for advancing human factors and HRI w.r.t trust in variable autonomy robotic systems. Of particular importance is developing a framework for measuring the different dimensions of trust, towards computational models of trust, capable of informing MI LoA switching policies. 






\bibliographystyle{IEEEtran}
\bibliography{IEEEabrv,refs}

\begin{thebibliography}{10}
\providecommand{\url}[1]{#1}
\csname url@rmstyle\endcsname
\providecommand{\newblock}{\relax}
\providecommand{\bibinfo}[2]{#2}
\providecommand\BIBentrySTDinterwordspacing{\spaceskip=0pt\relax}
\providecommand\BIBentryALTinterwordstretchfactor{4}
\providecommand\BIBentryALTinterwordspacing{\spaceskip=\fontdimen2\font plus
\BIBentryALTinterwordstretchfactor\fontdimen3\font minus
  \fontdimen4\font\relax}
\providecommand\BIBforeignlanguage[2]{{%
\expandafter\ifx\csname l@#1\endcsname\relax
\typeout{** WARNING: IEEEtran.bst: No hyphenation pattern has been}%
\typeout{** loaded for the language `#1'. Using the pattern for}%
\typeout{** the default language instead.}%
\else
\language=\csname l@#1\endcsname
\fi
#2}}

\bibitem{Bainbridge1983}
L.~Bainbridge, ``{Ironies of automation},'' \emph{Automatica}, vol.~19, no.~6,
  pp. 775--779, 1983.

\bibitem{Parasuraman2000}
R.~Parasuraman, T.~B. Sheridan, and C.~D. Wickens, ``{A model for types and
  levels of human interaction with automation.}'' \emph{IEEE transactions on
  systems, man, and cybernetics. Part A, Systems and humans}, vol.~30, no.~3,
  pp. 286--297, 2000.

\bibitem{DeVisser2018}
E.~J. de~Visser, R.~Pak, and T.~H. Shaw, ``{From ‘automation' to
  ‘autonomy': the importance of trust repair in human–machine
  interaction},'' \emph{Ergonomics}, vol.~61, no.~10, pp. 1409--1427, 2018.

\bibitem{Chiou2016IROS_HI}
M.~Chiou, R.~Stolkin, G.~Bieksaite, N.~Hawes, K.~L. Shapiro, and T.~S.
  Harrison, ``{Experimental analysis of a variable autonomy framework for
  controlling a remotely operating mobile robot},'' in \emph{2016 IEEE/RSJ
  International Conference on Intelligent Robots and Systems (IROS)}, 2016, pp.
  3581--3588.

\bibitem{Chiou2021_arXiv}
M.~Chiou, N.~Hawes, and R.~Stolkin, ``{Mixed-Initiative variable autonomy for
  remotely operated mobile robots},'' \emph{ACM Transactions on Human-Robot
  Interaction}, vol.~10, no.~4, 2021.

\bibitem{Chen2014}
J.~Y.~C. Chen and M.~J. Barnes, ``{Human–Agent Teaming for Multirobot
  Control: A Review of Human Factors Issues},'' \emph{IEEE Transactions on
  Human-Machine Systems}, vol.~44, no.~1, pp. 13--29, 2014.

\bibitem{Lee2004}
J.~D. Lee and K.~A. See, ``{Trust in Automation: Designing for Appropriate
  Reliance},'' \emph{Human Factors: The journal of the human factors and
  ergonomics society}, vol.~46, no.~50, pp. 50--80, 2004.

\bibitem{Hoff2015}
K.~A. Hoff and M.~Bashir, ``{Trust in automation: Integrating empirical
  evidence on factors that influence trust},'' \emph{Human Factors}, vol.~57,
  no.~3, pp. 407--434, 2015.

\bibitem{Schaefer2016}
K.~E. Schaefer, J.~Y. Chen, J.~L. Szalma, and P.~A. Hancock, ``{A Meta-Analysis
  of Factors Influencing the Development of Trust in Automation: Implications
  for Understanding Autonomy in Future Systems},'' \emph{Human Factors},
  vol.~58, no.~3, pp. 377--400, 2016.

\bibitem{Hancock2011}
P.~A. Hancock, D.~R. Billings, K.~E. Schaefer, J.~Y. Chen, E.~J. {De Visser},
  and R.~Parasuraman, ``{A meta-analysis of factors affecting trust in
  human-robot interaction},'' \emph{Human Factors}, vol.~53, no.~5, pp.
  517--527, 2011.

\bibitem{Matthews2020}
G.~Matthews, J.~Lin, A.~R. Panganiban, and M.~D. Long, ``{Individual
  Differences in Trust in Autonomous Robots: Implications for Transparency},''
  \emph{IEEE Transactions on Human-Machine Systems}, vol.~50, no.~3, pp.
  234--244, 2020.

\bibitem{Chien2020}
S.-Y. Chien, M.~Lewis, K.~Sycara, A.~Kumru, and J.~S. Liu, ``{Influence of
  Culture, Transparency, Trust, and Degree of Automation on Automation Use},''
  \emph{IEEE Transactions on Human-Machine Systems}, vol.~50, no.~3, pp.
  205--214, 2020.

\bibitem{Wright2020}
J.~L. Wright, J.~Y. Chen, and S.~G. Lakhmani, ``{Agent Transparency and
  Reliability in Human-Robot Interaction: The Influence on User Confidence and
  Perceived Reliability},'' \emph{IEEE Transactions on Human-Machine Systems},
  vol.~50, no.~3, pp. 254--263, 2020.

\bibitem{Yang2017}
X.~J. Yang, V.~V. Unhelkar, K.~Li, and J.~A. Shah, ``{Evaluating Effects of
  User Experience and System Transparency on Trust in Automation},'' in
  \emph{ACM/IEEE International Conference on Human-Robot Interaction (HRI)},
  2017, pp. 408--416.

\bibitem{Desai2012}
M.~Desai, M.~Medvedev, M.~V{\'{a}}zquez, S.~McSheehy, S.~Gadea-Omelchenko,
  C.~Bruggeman, A.~Steinfeld, and H.~Yanco, ``{Effects of changing reliability
  on trust of robot systems},'' in \emph{ACM/IEEE International Conference on
  Human-Robot Interaction (HRI)}, 2012, pp. 73--80.

\bibitem{Nam2020}
\BIBentryALTinterwordspacing
C.~Nam, P.~Walker, H.~Li, M.~Lewis, and K.~Sycara, ``{Models of Trust in Human
  Control of Swarms With Varied Levels of Autonomy},'' \emph{IEEE Transactions
  on Human-Machine Systems}, vol.~50, no.~3, pp. 194--204, 2020. [Online].
  Available: \url{https://ieeexplore.ieee.org/document/8651317/}
\BIBentrySTDinterwordspacing

\bibitem{Wang2016}
N.~{Wang}, D.~V. {Pynadath}, and S.~G. {Hill}, ``Trust calibration within a
  human-robot team: Comparing automatically generated explanations,'' in
  \emph{ACM/IEEE International Conference on Human-Robot Interaction (HRI)},
  2016, pp. 109--116.

\bibitem{Soh2018}
H.~Soh, Y.~Xie, M.~Chen, and D.~Hsu, ``{Multi-Task Trust Transfer for
  Human-Robot Interaction},'' \emph{International Journal of Robotics
  Research}, vol.~39, no. 2-3, pp. 233--249, 2018.

\bibitem{Rotter1966}
J.~B. Rotter, ``{Generalized expectancies of internal versus external control
  of reinforcements},'' \emph{Psychological monographs: General and applied},
  vol.~80, no.~1, pp. 1--28, 1966.

\bibitem{Furnham1993}
A.~Furnham and H.~Steele, ``{Measuring locus of control: A critique of general,
  children's, health‐ and work‐related locus of control questionnaires},''
  \emph{British Journal of Psychology}, vol.~84, no.~4, pp. 443--479, 1993.

\bibitem{Takayama2011}
L.~Takayama, E.~Marder-Eppstein, H.~Harris, and J.~M. Beer, ``{Assisted driving
  of a mobile remote presence system: System design and controlled user
  evaluation},'' in \emph{IEEE International Conference on Robotics and
  Automation}, 2011, pp. 1883--1889.

\bibitem{Acharya2018}
U.~Acharya, S.~Kunde, L.~Hall, B.~A. Duncan, and J.~M. Bradley, ``{Inference of
  user qualities in shared control},'' in \emph{IEEE International Conference
  on Robotics and Automation (ICRA)}, 2018, pp. 588--595.

\bibitem{jiang2015mixed}
S.~Jiang and R.~C. Arkin, ``Mixed-initiative human-robot interaction:
  definition, taxonomy, and survey,'' in \emph{2015 IEEE International
  Conference on Systems, Man, and Cybernetics}.\hskip 1em plus 0.5em minus
  0.4em\relax IEEE, 2015, pp. 954--961.

\bibitem{Chiou2019_SMC_learning_effects}
M.~Chiou, M.~Talha, and R.~Stolkin, ``{Learning effects in variable autonomy
  human-robot systems: how much training is enough?}'' in \emph{IEEE
  International Conference on Systems, Man and Cybernetics (SMC)}, 2019, pp.
  720--727.

\bibitem{Marder-Eppstein2010}
E.~Marder-Eppstein, E.~Berger, T.~Foote, B.~Gerkey, and K.~Konolige, ``{The
  office marathon: Robust navigation in an indoor office environment},'' in
  \emph{IEEE International Conference on Robotics and Automation (ICRA)}, 2010,
  pp. 300--307.

\bibitem{Mathot2012OpenSesame}
S.~Math{\^{o}}t, D.~Schreij, and J.~Theeuwes, ``{OpenSesame: An open-source,
  graphical experiment builder for the social sciences},'' \emph{Behavior
  Research Methods}, vol.~44, no.~2, pp. 314--324, 2012.

\bibitem{Ganis2015}
G.~Ganis and R.~Kievit, ``{A New Set of Three-Dimensional Shapes for
  Investigating Mental Rotation Processes : Validation Data and Stimulus
  Set},'' \emph{Journal of Open Psychology Data}, vol.~3, no.~1, 2015.

\bibitem{Murphy2004}
R.~R. Murphy, ``{Human-Robot Interaction in Rescue Robotics},'' \emph{IEEE
  Transactions on Systems, Man and Cybernetics, Part C (Applications and
  Reviews)}, vol.~34, no.~2, pp. 138--153, 2004.

\bibitem{Hoffman2013}
G.~Hoffman, ``{Evaluating Fluency in Human-Robot Collaboration},'' in
  \emph{International conference on human-robot interaction (HRI), workshop on
  human robot collaboration}, 2013.

\bibitem{Nikolaidis2017}
S.~Nikolaidis, D.~Hsu, and S.~Srinivasa, ``{Human-robot mutual adaptation in
  collaborative tasks: Models and experiments},'' \emph{The International
  Journal of Robotics Research}, vol.~36, no. 5-7, pp. 618--634, 2017.

\bibitem{Duttweiler1984}
P.~C. Duttweiler, ``{The Internal Control Index: A Newly Developed Measure of
  Locus of Control},'' \emph{Educational and Psychological Measurement},
  vol.~44, no.~2, pp. 209--221, 1984.

\bibitem{Meyers1988}
L.~S. Meyers and D.~T. Wong, ``{Validation of a new test of locus of control:
  The internal control index},'' \emph{Educational and Psychological
  Measurement}, vol.~48, no.~3, pp. 753--761, 1988.

\bibitem{Maltby1996}
J.~Maltby and C.~D. Cope, ``{Reliability estimates of the internal control
  index among UK samples},'' \emph{Psychological Reports}, vol.~79, no.~2, pp.
  595--598, 1996.

\bibitem{Sharek2011}
D.~Sharek, ``{A Useable, Online NASA-TLX Tool},'' \emph{Proceedings of the
  Human Factors and Ergonomics Society Annual Meeting}, vol.~55, no.~1, pp.
  1375--1379, 2011.

\bibitem{Twenge2004}
J.~M. Twenge, L.~Zhang, and C.~Im, ``{It's beyond my control: A cross-temporal
  meta-analysis of increasing externality in locus of control, 1960-2002},''
  \emph{Personality and Social Psychology Review}, vol.~8, no.~3, pp. 308--319,
  2004.

\bibitem{Chancey2017b}
E.~T. Chancey, J.~P. Bliss, Y.~Yamani, and H.~A. Handley, ``{Trust and the
  Compliance-Reliance Paradigm: The Effects of Risk, Error Bias, and
  Reliability on Trust and Dependence},'' \emph{Human Factors}, vol.~59, no.~3,
  pp. 333--345, 2017.

\bibitem{Chen2018}
M.~Chen, S.~Nikolaidis, H.~Soh, D.~Hsu, and S.~Srinivasa, ``{Planning with
  Trust for Human-Robot Collaboration},'' in \emph{ACM/IEEE International
  Conference on Human-Robot Interaction}, 2018, pp. 307--315.

\end{thebibliography}

\end{document}